# Challenges of engineering safe and secure highly automated vehicles
## Whitepaper

Nadja Marko[1], Eike Möhlmann[2], Dejan Ničković[3], Jürgen Niehaus[4], Peter Priller[5] and Martijn Rooker[6]

[1] *Virtual Vehicle, Graz, Austria,* [2] *OFFIS, Oldenburg, Germany,* [3] *AIT Austrian Institute of Technology, Vienna, Austria,* [4] *SafeTRANS, Oldenburg, Germany,* [5] *AVL, Graz, Austria,* [6] *TTTech, Vienna, Austria*

## Contents

## Abstract

After more than a decade of intense focus on automated vehicles, we are still facing huge challenges for the vision of fully autonomuous driving to become a reality. The same 'disillusionment' is true in many other domains, in which autonomuous Cyber-Physical Systems (CPS) could considerably help to overcome societal challenges and be highly beneficial to society and individuals.

Taking the automotive domain – i.e., highly automated vehicles (HAV) – as an example, this paper sets out to summarize the major challenges that are still to overcome for achieving safe, secure, reliable and trustworthy highly automated resp. autonomous CPS. We constrain ourselves to technical challenges, acknowledging the importance of (legal) regulations, certification, standardization, ethics, and societal acceptance, to name but a few, without delving deeper into them as this is beyond the scope of this paper.

Four challenges have been identified as being the main obstacles to realizing HAV: Realization of continuous, post-deployment systems improvement, handling of uncertainties and incomplete information, verification of HAV with machine learning components, and prediction. Each of these challenges is described in detail, including sub-challenges and, where appropriate, possible approaches to overcome them.

By working together in a common effort between industry and academy and focussing on these challenges, the authors hope to contribute to overcome the 'disillusionment' for realizing HAV.

Date: October 2020

Challenges of engineering safe and secure highly automated vehicles

# Introduction

After more than a decade of intense focus on automated driving by the industry (both automotive and IT), huge investments and bold announcements[1,2], Gartner put SAE[3] level 4 cars on the declining branch, beyond their peak, on the hype cycle[4]. Self-driving cars are now in the "valley of disillusionment", and it is not clear if current approaches will suffice to climb the slope of enlightenment in time before public confidence is lost. Each fatality with an automated vehicle involved [5, 6] shakes confidence in achieving the declared goal of drastically improving road safety. Examples for highly automated vehicles in operation exist (e.g. shuttles[7]), but are typically constrained to pre-defined routes and/or limited to low-speed travel only.

In order to leave the valley of tears behind, essentially two aspects are needed:

a) We need to change the way highly automated vehicles (HAV) are developed and tested. For this, existing design and testing processes need to be extended to processes covering the whole life cycle of an HAV. These extended processes enable usage of (performance and other) data collected in the field for continuous engineering of HAV. Together with corresponding monitoring and update methods as well as in conjunction with systems properties like fail-awareness and fail-operationality, these processes allow for validation of safety cases that are much more appropriate to the dynamic nature of HAV.

b) We need to integrate continuous design into homologation/certification processes that are (i) fit to their purpose; obviously, a system passing homologation should be guaranteed to have the required safety, security, reliability, and other qualities, (ii) accepted by regulation authorities, i.e. homologation should be performed in a way that all safety arguments are understandable and verifiable by a third party, and (iii) accepted by the public / the end user of these systems, i.e., the public must be able to trust that these systems have the desired qualities.

The main focus of this paper is on a) above; however, the need to embed the new and extended design processes into homologation and certification frameworks presents restrictions on the kind of solutions and technologies that can be used to solve the main challenges in a), and must therefore always be considered. Second, this requirement immediately points to the importance of open standards for all aspects of the homologation process, including the life cycle spanning design and testing processes. Open standards are needed, regardless of whether homologation for e.g., highly automated cars follows self-certification concepts like e.g., in the US, or third-party certification like e.g., in the EU, in order to enable the necessary fitness for purpose, acceptance by regulation authorities, and trust by end-users.

There are already quite extensive roadmaps on the topic of highly autonomous cyber-physical systems[8] resp. autonomous vehicles as well as corresponding strategy documents[9]. Typically, these are either very elaborated, spanning the whole CPS domain and taking a general approach to CPS development and validation, without tailoring to specific domains, like automotive; or they describe challenges and R&D strategies in great detail, making it sometimes difficult to see the key elements. This paper, in contrast, is intended to identify **the essential technical challenges** which need to be solved for safe, reliable and secure automated vehicles. The main focus is on safety, considering (but not explicitly elaborating) the impact of security on safety. Each challenge is described briefly, rationalizing why we believe these challenges are relevant. The scope of this paper is deliberately limited to a concise elaboration of the challenges. However, the reader is encouraged to read the more detailed roadmaps and strategy documents to get a deeper view on the topic. In addition, although we acknowledge the fact that there are already a number of public funded research activities as well as industry initiatives[10] addressing these challenges in different combinations, we still feel that (a) because of the enormity of the tasks to solve these can only provide part of the solutions and (b) a concise, yet holistic presentation of the challenges is needed in order to grasp the enormity and urgency of the problem.

---

[1] Musk predicts that Tesla will have autonomous robotaxis without drivers in some U.S. markets in 2020. https://www.reuters.com/article/us-tesla-autonomous-factbox-idUSKCN1RY0QY

[2] Nvidia & Audi Set a 2020 Timeframe for Level 4 Autonomous Cars: https://www.sdxcentral.com/articles/news/nvidia-audi-set-2020-timeframe-level-4-autonomous-cars/2017/01/

[3] SAE J3016, see https://www.sae.org/standards/content/j3016_201401/

[4] https://www.gartner.com/smarterwithgartner/5-trends-appear-on-the-gartner-hype-cycle-for-emerging-technologies-2019/

[5] Tesla may have been on Autopilot in California crash which killed two https://www.theguardian.com/technology/2020/jan/01/tesla-autopilot-california-crash-two-deaths

[6] https://www.tesladeaths.com/

[7] E.g., Shared Personalised Automated Connected vEhicles (SPACE), https://space.uitp.org/

[8] E.g., ECSEL Joint Undertaking: Strategic Research Agenda, 2020
Peter Heidl, Werner Damm. Highly Automated Systems: Test, Safety, and Development Processes - Research Challenges and Recommendations of Actions - Management Summary. 2017;
Lemmer, K. (Hrsg.): Neue autoMobilität II. Kooperativer Straßenverkehr und intelligente Verkehrssteuerung für die Mobilität der Zukunft (acatech STUDIE), München: utzverlag GmbH 2019
SafeTRANS (Hrsg.). Safety, Security, and Certifiability of Future Man-Machine Systems - Positionspapier (Draft-Version). Oldenburg. 2019.

[9] E.g., SaFAD group (Daimler, BMW, Audi, VW, and others). "Safety first for automated driving"., 2019

As pointed out above, the focus of our consideration is a life cycle process for highly automated vehicles, based on the concept of continuous engineering. We do believe however, that many of the aspects discussed are also relevant for partially automated vehicles and can be transferred to highly automated cyber-physical systems (ACPS) in other domains as well. It combines design, verification and validation (V&V) and operation in a continuous flow keeping safety, reliability, security and resilience aspects in view. For sake of clarity, the paper is limited to the technical development, not covering a broader discussion of societal aspects of automated and autonomous driving.

The following identification and description of challenges shall contribute to a targeted orientation of research towards required solutions. The authors of this paper expect that this focus is needed to "get out of the ditch", pictured in the Gartner cycle, to achieve mature automated vehicles, which is of critical importance to the future of autonomous system dependability overall. Mastering these challenges will bring the lead in safe, reliable and secure automated vehicles, with convincing means of proof, earning sustained trust in our society.

## Challenge 1: Continuous, post-deployment systems improvement

While vehicle manufacturers have a lot of know-how in developing conventional vehicles with limited automation, full-fledged experience in developing (fully) automated vehicles (L3 – L5) is lacking, as this is a rather new field of application. Of course, autonomous shuttles already exist, but their operational context (Operational Design Domain – ODD) is limited, e.g. they only follow a proven track.

The development of automated systems and especially vehicles needs adapted and improved development processes. Whereas the development of non-automated vehicles was nearly finished after production, highly automated vehicles, but also partly automated vehicles, must be continuously improved. The need for the continuous improvement results from a) new requirements (e.g., new traffic signs, new traffic rules), b) fixing errors or vulnerabilities (e.g. new attacks), c) improving the vehicle performance or d) increasing the comfort to keep vehicles comfortable and trustworthy (cp. sub challenge 4.4). New requirements permanently come up and must be fulfilled to ensure customer satisfaction and to avoid critical situations. Hence, in addition to developing new vehicles, vehicle manufacturers have to continuously adapt existing vehicles which are in operation and consider this in the development process. Further, upcoming standards, like the ISO/SAE 21434 cybersecurity standard or the UN ECE regulation on updates[11], require update processes, which must be considered in vehicle development already in the near future. This makes a close cooperation between partners in the whole supply chain necessary. Figure 1 shows a simplified scheme for the continuous improvement.

---

[10] Like the ENABLE-S3 and the ArchitectECA2030 projects within the ECSEL Joint Undertaking, or the PEGASUS and the "KI project family" in Germany, to name but a few.

[11] United Nations Economic Commission for Europe (UN ECE), World Forum on Harmonization of Vehicle Regulations (WG.29), Working Party on Automated/Autonomous and Connected Vehicles (GRVA), Task Force on Cyber Security and (OTA) software updates (CS/OTA).
https://wiki.unece.org/pages/viewpage.action?pageId=40829521

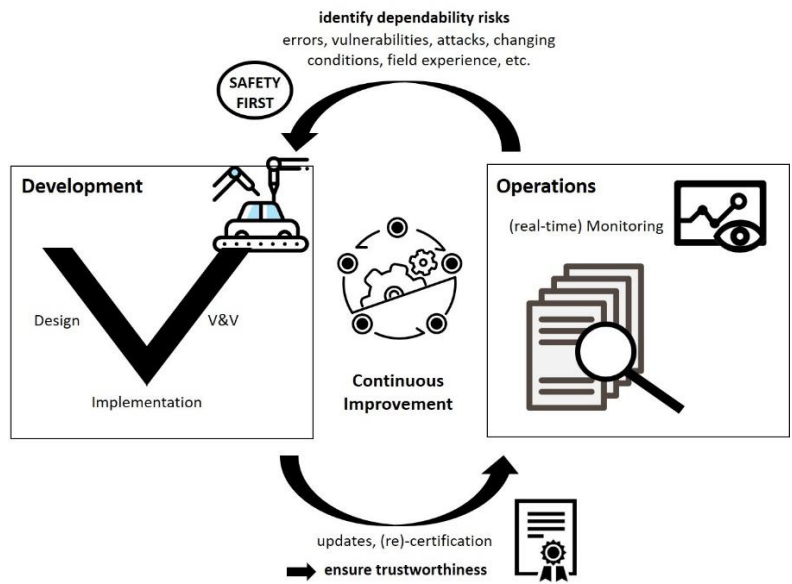

*Figure 1 – Simplified scheme for continuous improvement*

| Sub Challenge 1.1 |
|---|
| Find system vulnerabilities, erroneous and missing functionality during operation |

In order to be able to improve vehicles during operation, it must be known which functionalities and components have to be improved and how to improve them. Advanced post-deployment observation is required, and system developers and engineers must develop vehicles with (self)monitoring components in addition to vehicle functionalities. These monitoring components must provide reliable information about the status of vehicle's safety, reliability and security. On the one hand, this data is needed in real-time during driving to make safe driving decisions. On the other hand, it is needed to analyze functions and components offline. Additionally, offline analysis should support finding vulnerabilities and faults to be able to make improvements. Finally, it should also support engineering activities to improve the engineering process, for example, extracting real-life traffic scenarios from monitored data which can be used to efficiently test automated driving functions or to refine functional requirements. This monitoring and data collection can be integrated in the general big-data collection techniques modern vehicles will very likely have anyway.

| Sub Challenge 1.2 |
|---|
| Fast improvement of automated vehicles in operation |

If it is determined that components are vulnerable or unsafe, immediate action must be taken. First and foremost, vehicles have to go in a safe fallback state, in which the critical functions are not used anymore or are only used in situations which are uncritical for the vehicle (fail-operational, fail-safe). The vehicle has to provide (a) the needed safety mechanisms and (b) send information to the OEM for improvement. The continuous improvement is related to (b). A fast improvement of vehicles is essential, as time plays a major role for customer satisfaction and for reduction of further risks from the same cause. If an automated function or even automated driving is not possible for a long time, customers will get annoyed and trustworthiness will be compromised. Hence, a fast improvement is crucial to avoid massive impact.

In a short time, vehicle manufacturers have to analyze monitored data to find the error (cp. challenge 1.3) and find out which vehicles are affected and have to be improved. Although there will be pressure to have relatively homogeneous fleets, there will be many system configurations and variants as hardware changes may occur and software updates will be needed. Appropriate methods have to be applied to quickly identify which configurations and variants have to be improved. In addition to fixing the vulnerable functionality, the impact of change on other components must be analyzed. It could be the case, that related functions have to be adapted as well, in order to avoid errors. Hence, traceability and automation, e.g. continuous integration, are essential for fulfilling the timing requirements in automated vehicle development. After the improvement of all concerned functions, individual systems and the vehicle have to be tested again, see challenge 1.4.

| Sub Challenge 1.3 |
| --- |
| Identify, filter and process relevant data from post-deployment observation |

The necessary monitoring components produce tons of data sets what makes it difficult to find out which data is important and should be analyzed further. Methods are needed to efficiently identify, filter, and process relevant data to improve functions and ensure safety of the vehicle. For this offline analysis, which is essential for the continuous improvement of HAV, developers have to filter relevant data, find vulnerabilities and analyze possible risks that influence safety of vehicles. These offline analysis actions have to be done continuously based on monitored data from the vehicle (e.g. error reports, newly detected scenarios or performance issues). Therefore, data relations between different components and functions have to be analyzed. Relevant information to identify critical situations arise often from various components. Due to the dynamics of real-world situations combined with the complexity of automated vehicles, this is a challenging task. Appropriate methods and online as well as offline software functions can help to facilitate the post-deployment analysis, which is done offline. Hence, these functions should be investigated further in order to save development time and therefore costs.

| Sub Challenge 1.4 |
| --- |
| Perform physical in-field tests for systems in operation (not available on-site) |

Testing updated software functions and components available on-site is time-consuming, also considering the exceeding number of system variants, which need to be verified and validated. With existing advanced methods and tools this activity can be supported. Remote diagnostics and monitoring functions can help to monitor and check the behavior of the updated functionality while in operation. However, testing systems, which are not available on-site, is hard. In-field tests also include vehicle-in-the-loop tests, which are not possible when systems are already deployed and not available anymore in the lab. Having all variants of vehicles available on-site is also not an option. Most probably, common variants will be physically present at vehicle manufacturer's site which allows also physical testing. However, other vehicle variants have to be tested virtually. Research on virtual representations of systems must be extended to provide precise enough digital twins that allow virtual V&V in addition to physical V&V. This might require a sophisticated monitoring concept to get field data to improve digital twins and also the selection of test cases.
In addition to these challenges, other related aspects must be considered that are beyond the scope of this work but are important as well. This includes:

- data protection (privacy, GDPR)
- communication and OTA Updates (addressed by the UN ECE regulation)
- Hardware changes
- Common/standardized interfaces

Novel activities and aligned engineering require research on advanced methods and tools to enable an efficient system improvement after deployment to avoid critical situations in traffic. New requirements demand for innovative systems engineering techniques and processes that support traceability, variability, configurability, modularity, virtualization and automation for development and the operational phase. Monitoring components have to become an inherent part in vehicles to support finding vulnerabilities and risks. Moreover, advanced monitoring functions are essential to be able to learn from vehicles in the field. Finding appropriate test cases from experiences on the road and realistic virtualization of components (e.g. digital twins) are examples for such required functionalities.
Handling post-deployment cases will become an extensive and challenging task. Beside the described technical challenges, vehicles must be kept trustworthy during evolution of systems. To be trustworthy, a kind of (re)certification must be considered as there can be side-effects on already proven functions when fixing an error or adding new/adapted vehicle functionalities. Re-certification must ensure that the vehicle is still safe, and homologation is valid, which is often a matter of costs.

# Challenge 2: Handling of uncertainties and incomplete information

Handling uncertainty and incomplete information stand out as major obstacles for developing, testing and operating HAV (and other ACPS). In these systems, **various sources of uncertainty and incomplete information** exist.

| Sub Challenge 2.1 |
|---|
| Statistical demonstration of system safety and a positive risk balance without driver interaction |

**First** of all, due to the complexity of the context in which HAV operate – consisting of the physical environment, static and dynamic objects in this environment, including other technical systems, humans, animals, different weather conditions, etc. – it is inherently impossible to define the ODD (Operational Design Domain[12]) of a system under development completely – i.e., in a way that it describes all possible situations that the HAV needs to cope with in reality. Defining such ODDs involves identifying the *relevant part of the context*, i.e., all environmental conditions and objects that the system under development needs to be able to cope and to interact with in a safe and secure way. For comparatively simple environments, identifying the relevant part of the context can be done with high confidence: For cars driving automatically on highways, for example, all typical environmental conditions, all typically occurring objects and their typical behavior are known. However, even in these simple environments, there always remains a small chance that something new and unforeseen will be encountered by the vehicle during operation in reality, which the designers failed to take into account when identifying the relevant context: Extreme weather conditions or objects legally not allowed on highways – e.g. cyclists or pedestrians – are typical examples of conditions that could be overlooked (cf. Figure 2).

However, the root of the problem goes deeper: literally *anything physically possible* can happen in reality, so any and all partitions of the context in a 'relevant part' and an 'non-relevant' part are inherently wrong. On the other hand, for defining ODDs – and therefore boundaries in which the system needs to operate safely – this partitioning is needed in order to have a finite specification for the ODD. Thus, even when operating in simple environments, HAV need to be able to cope with artefacts and conditions that are unknown to them – i.e., against which they have not been tested. For more complex environments, for example in urban traffic, the chance of encountering unknown situations, objects and conditions in reality increases considerably.

Even within a given ODD, the number of possible properties of the environment, the number of possible behaviors of dynamic objects in this environment, and the complexity of the behavior of the HAV itself are that huge, that exhaustive testing of all combinations of these conditions is prohibitive, both time- and cost-wise. Therefore, testing needs to focus on typical combinations of conditions, behaviors and traffic situations, i.e., on typical use-cases or scenarios. Thus, there are always situations (scenarios) (a) that can happen in reality, even if only with very low probability, but (b) against which the HAV has not been tested. From a HAV point of view, these are called *unknown scenarios*.

---

[12] An ODD comprises the "operating conditions under which a given driving automation system or feature thereof is specifically designed to function, including, but not limited to, environmental, geographical, and time-of-day restrictions, and/or the requisite presence or absence of certain traffic or roadway characteristics." (in: Surface Vehicle Recommended Practice — Taxonomy and Definitions for Terms Related to Driving Automation Systems for On-Road Motor Vehicles. SAE:J3016, 2018)

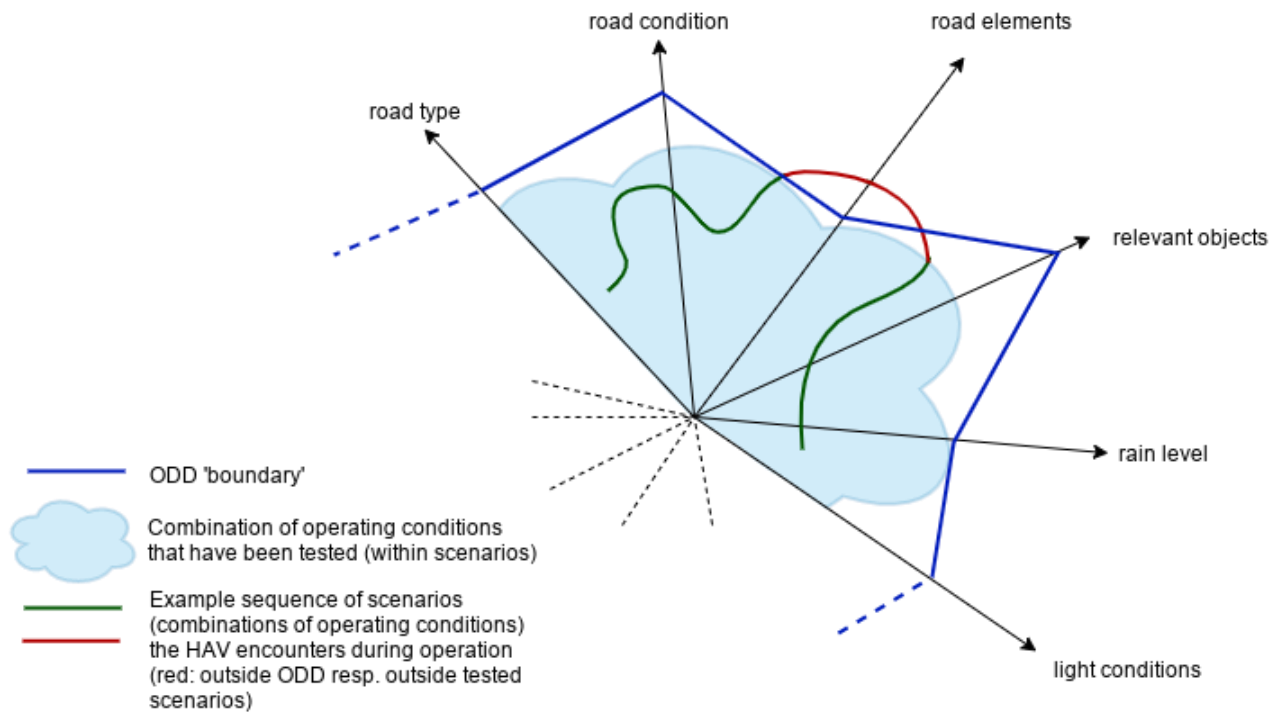

*Figure 2 – Simplified, flattened scheme of a multi-dimensional ODD definition, with scenarios tested during development and encountered during operation*

In traditional vehicles, i.e., SAE L0-2 systems, the human operator (driver) is responsible for handling the vehicle even when it operates outside of its ODD or when it operates in unknown scenarios. For higher levels of automation and autonomy, this responsibility rests with the HAV itself, so much more care needs to be taken to show that the system itself is able to handle any and all encountered situations in a safe way.

The first major challenge here is therefore to devise methods for statistical demonstration of safety assurance, to sufficiently reduce the residual risk of operating the automated vehicle and achieve a positive risk balance (wrt. human controlled vehicles).
 These methods need to
- enable Engineers to
    - define ODDs such that the HAV will with high statistical probability be able to always operate within the ODD,
    - and, within a given ODD, identify scenarios for testing the HAV that cover a sufficiently large part of that ODD, again such that the HAV will with high probability in reality only encounter these scenarios
- enable the HAV to
    - detect the – rare – cases in which it operates outside of its ODD or encounters situation not covered by a scenario against which it has been tested
    - cope with these in a safe way

To be more precise, within this sub challenge we need methods and supporting tools that
- enable Engineers during the Design and Test phase to
    - determine the **part of a context that is relevant**, i.e. should be included in the ODD, with sufficiently high statistical confidence
    - determine **ODD coverage of scenarios** with sufficiently high statistical confidence
    - determine **the residual risk** of operating an automated vehicle within a given ODD and tested with a given set of scenarios with sufficiently high statistical confidence; in addition, Engineers should be enabled to determine, which deviations from the ODD have the highest impact on system safety.
- enable the HAV (during (safe) operation) to

- determine at all times that it is still **operating within its ODD (including detection of (foreseeable) misuse).**
- determine at all times that it is **operating within scenarios** against which it has been **tested** (or within scenarios that are 'equivalent' to the tested ones according to a 'safe equivalence relation').
- be able to **degrade gracefully to a safe minimal functionality** if it detects operation outside of its ODD or outside of tested scenarios. This includes the ability to recover to full functionality if the reason(s) for degradation cease to exist.
- be able to detect, record and report all **relevant data of unknown/untested scenarios** which it might encounter.

The first three bullets or the operation phase imply HAV to be fail-aware and, at least for safety-critical functions, implement fail-safe or rather fail-operational behavior.

Unknown/untested scenarios detected and reported by the automated vehicle during operation should be analyzed (potentially using AI techniques, cf. challenge 3) for their potential to being used to develop and test updates and new versions of the automated vehicles' functionality (cf. challenge 1). Detecting and reporting these unknown scenarios can very well be combined with a general data collection strategy to derive test scenarios from real world data, and, obviously and just like this general data collection strategy, must be made in a way protecting privacy rights of users of HAV, i.e., in accordance with GDPR and similar regulations.

| Sub Challenge 2.2 |
|---|
| Make safe decisions based on incomplete and inaccurate data |

**A second major source of uncertainty and incomplete information** arises during the nominal operation of automated vehicles:

- Sensors are only able to sense with a certain accuracy. This accuracy depends upon various environmental conditions. Filtering techniques for sensor data can distinguish noise from real data, but only with certain, sensor dependent accuracy. Combining different sensors (e.g. lidar, radar or camera) using advanced forms of sensor fusion techniques is used to increase the total accuracy with which objects can be detected; still, this is not sufficient to guarantee that all objects in the environment are detected at all times with absolute certainty. While non-detection of existing objects has obvious safety implications, note that false positives, i.e., 'detecting' objects that are actually not there, severely influences performance and thus user acceptance of the system. Summarizing, object detection and classification, and deriving a 'mental map' or 'internal world model', is inherently uncertain.
- In addition, the 'mental map' or 'internal world model' of the system is a data structure that has been specified at design time by the designers. This model is meant to contain representations of all detected objects as well as their relevant properties and relations amongst each other. With a similar reasoning as for challenge 2.1., it cannot be guaranteed that this model is complete with respect to all needed object properties and relations, thus forming another source of potentially incomplete information.
- For dynamic (i.e. potentially moving) objects in the environment, it is typically not sufficient to know their current location and trajectories. Instead, the automated vehicle needs to predict the future movement of these objects, including changes in their trajectory, to be able to safely plan its own trajectory. In general, for any encountered situation/scenario, the future evolution of this situation needs to be predicted. This prediction is eased if other objects (e.g., other cars or other technical systems) are able to communicate and willing to cooperate. However, since many objects in the environment will not be connected (or might be unwilling or unable to cooperate), this prediction has to rely on statistical information in general, thus creating a second source of uncertainty. See Challenge 4 for a more detailed consideration of challenges wrt. prediction.
- Last, but not least, automated vehicles often depend upon external data-sources for their decision making, i.e., on high accuracy maps, on information sent by other systems about their intended behavior, on weather forecast or other situational data received from 'the cloud'. This data may be inaccurate dependent on how and by whom it was collected and processed. In addition, malicious intent (by 'hackers' or by other systems that want to achieve unfair advantages) may cause this data to be inaccurate and thus impair decisions derived from it.

The challenge to be solved here is therefore to enable the HAV to make safe decisions based on incomplete and inaccurate data.

This sub-challenge can be further decomposed into:
- Determine **statistical confidence and accuracy of sensor data** under all (relevant) environmental conditions (including overall confidence and accuracy levels of sensor fusion techniques)
- Determine **statistical confidence and accuracy of prediction functions** (i.e. of functions that predict the future movement and behavior of dynamic objects in the environment)
- Expand prediction functions to not only predict the 'average case', but to simultaneously **predict all safety-relevant cases** and assign a statistical confidence to each of them.
- Determine 'trustworthiness' and **statistical confidence for all external data sources** (taking cyberattacks into account)
- Devise decision algorithms that take into account uncertain and incomplete information and their confidence levels, and that are able to take decisions that always lead to (a) safe behavior of the system, and (b) to high performance and availability of the system. Point (a) means that the system should have a sufficiently low, acceptable risk of taking a wrong decision, while (b) ensures that the system is not too cautious but performs as expected.

## Challenge 3: Verification of Highly Automated Vehicles with Machine Learning Components

Components based on machine learning (ML) are at the core of highly automated vehicles. They pose a significant challenge to trust in this emerging technology and there is an urgent need to develop novel methodologies that address verification of highly automated vehicles in presence of ML components.

| Sub Challenge 3.1 |
|---|
| Ensure that the ML components are robust to input perturbations. |

An important basic requirement for trustworthiness of ML components is their robustness to input perturbations. A robustness property indicates how effective the ML algorithm is to new independent inputs that are either noisy or similar to training inputs. For instance, there are successful experiments in fooling vision algorithms by only changing a single pixel in the picture[13].

The problem of robustness in ML algorithms can be at least partially addressed by adversarial learning[14], which explicitly trains the model using adversarial (malicious) examples. Adversarial learning can effectively detect non-robust behavior in ML algorithms but cannot be used to prove robustness. Proving robustness of ML methods can be tackled by the application of formal verification techniques. Scalability remains an issue in applying the formal verification to ML components of realistic size.

| Sub Challenge 3.2 |
|---|
| Precisely formulate high-level requirements for the ML components. |

Robustness to input perturbations is a low-level property that naturally admits a precise mathematical formulation. Formally specifying expected properties of ML components is crucial in facilitating their systematic and automated verification. However, capturing high-level application-dependent properties of machine learning components may be much less straight-forward. We note that ML algorithms in highly automated vehicles are typically used for perception and control modules. There is a fundamental difference between these two applications of ML. Data-driven control typically admits a precise mathematical formulation of its desired properties. This is not the case for the perception modules commonly used in many safety critical systems such as highly automated vehicles. For example, the Uber test vehicle in the Arizona 2018 accident failed to correctly identify the bicycle as an imminent collision until just before impact[15]. Intuitively, we would like the perception component to correctly detect all and only the elements that impact control decisions. Data used to train the perception module forms an implicit specification for the above requirement. It is important to validate that the training data is correct and complete (according to some appropriate notion of correctness and completeness) and that it does not induce bias in the perception component. We also note that neither this requirement nor its implicit specification based on the training data are not mathematically precise and it is not clear how can one formulate them using a formal and rigorous specification language.

---
[13] "Single pixel change fools AI programs". BBC News. 3 November 2017. Retrieved 12 February 2018
[14] Vorobeychik, Yevgeniy, and Murat Kantarcioglu. Adversarial machine learning. Morgan & Claypool Publishers, 2018.

| Sub Challenge 3.3 |
| --- |
| Develop effective methods for verification of ML components and system-level verification of highly automated vehicles that integrate ML components. |

Verification has tremendously matured over the last few decades and successfully used on daily basis to evaluate correctness and performance of digital and analog hardware and complex embedded systems. Unfortunately, the classical verification and testing methods cannot be directly applied to many ML components. This is because many ML algorithms exhibit complex internal behaviors that are not commonly seen for traditional verification. ML components also do not exist in isolation but are part of a larger system and interact with other, possibly non-AI-based sub-systems. For example, a misclassification of a pedestrian on the street in front of an automated vehicle can have a much more serious impact on its control algorithm than the misclassification of a house that is sufficiently far away from the street.
Hence, it is important to develop effective methods for both for verification of ML components in isolation and for system-level verification of ML-enabled HAV. Systematic black-box testing methods based on search-based testing and simulation-based sensitivity analysis can be used to reason about ML algorithms and their system-level integration. Many ML methods, such as Deep Neural Networks (DNN), can be formulated as sets of non-linear differential equations, admitting analysis using white-box verification methods.

| Sub Challenge 3.4 |
| --- |
| Make ML components transparent and explainable. |

Many ML algorithms, including the widely popular and successful deep learning methods are opaque and are typically used as black boxes. In safety-critical applications, in which decisions that can affect human lives are taken based on the outcomes of ML algorithms, this lack of transparency is a major problem. For instance, decisions made by a DNN-based data-driven controller are typically not comprehendible by the design engineer, because they cannot see the inner working mechanisms of the DNN. Achieving trustworthiness of ML components and enabling their certification requires the design and verification engineers, as well as regulatory bodies to understand and explain how the ML component works and how the autonomous decisions are taken.
There are two major and complementary directions in which the problem of transparency can be addressed. The first direction is to use transparent-by-construction ML algorithms. The second direction is to use explainable artificial intelligence methods that are based on sensitivity and impact analysis to identify parts of the input and/or system space that are responsible for each individual decision.

| Sub Challenge 3.5 |
| --- |
| Ensure that safety properties are preserved for evolvable HAV with continuous learning ML components. |

---

[15] Wakabayashi, Daisuke (March 19, 2018). "Self-Driving Uber Car Kills Pedestrian in Arizona, Where Robots Roam". The New York Times. Retrieved March 22, 2018.

Finally, a long term research and technology goal is to enable development of ML components that evolve over time, adapt their behavior to the newly discovered scenarios and learn new situations. For example, handling of the impact of infectious diseases like e.g., the Covid-19 pandemics, may require perception algorithms to be re-trained in order to take into account the new reality of pedestrians systematically wearing masks. The continuous learning can be applied to the ML algorithm in an offline fashion and then updated to the component or it can be applied online during the operation of the ML component. In both cases, it is a very challenging problem to make sure that the adapted ML component still guarantees safe operation.

One promising direction in ensuring safe adaptive and evolvable systems is to use safe reinforcement learning methods, in which the safety objectives are used to shape the reward function. Another promising direction is to allow the component to continuously learn and evolve, but to have a safety supervisor that monitors the component during its operation, detects any potential violations of the safety requirements and accordingly switches to a fail-safe mode when necessary.

## Challenge 4: Challenges in Prediction

Predicting the future behavior of objects in the environment of automated or autonomous systems is comparatively simple in environments which are under full control (e.g. labs or factories with specially trained humans). However, for future operational areas, such as automated or autonomous driving, farming, or naval transport, this is not the case. While the perception of the objects can be difficult, e.g., under bad weather conditions or occlusion, it is even more difficult to predict future behavior and future trajectories of objects, like humans, which have to be safe guarded. This is needed not only to prevent accidents but also for a successful operation.

| Sub Challenge 4.1 |
| --- |
| Predict and handle not only the nominal case (or most likely behavior) but also for the abnormal behavior. |

As already mentioned, demonstration of safety is a complicated task for systems with ML components. ML is often used for prediction of surrounding objects. Usually prediction requires behavioral models of the surrounding objects. Deriving good models of the environment (including the relevant objects) is a complex task, which usually is done by experts or learned from historic data. Hence, these models induce uncertainty and are biased towards nominal behavior e.g. due to the lack of data for abnormal or rule-violating behavior. Therefore, the degree of ensured safety depends on the range of possible behaviors considered by prediction and planning components and that a possible bias is taken into account by the safety mechanisms.

| Sub Challenge 4.2 |
| --- |
| Identification of relevant objects and relevant behavior. |

There is no complete description of all objects (and their possible shapes) for most ODDs of highly automated systems. Instead one focuses on *relevant objects*, i.e., objects that may influence the system or can be influenced by the system. But it is still difficult to enumerate all objects which are relevant in the surrounding of the considered system and often this is done exemplarily: cars, bike, pedestrians, wheelchairs, and so on; but when is the list sufficiently complete? For some objects — including those which are (a) mentioned in the safety requirements, (b) commonly observed, or (c) considered by domain experts — relevance is known a priori, but others that are rare or considered unimportant by humans may be very important for an automated system. This is especially, but not only, the case for systems with machine learning components.

In order to predict the behavior of the relevant objects, it is further needed to have a description of their relevant behavior. Obtaining a precise description of the relevant objects regarding their potential behavior is mostly impossible. Instead worst-case assumptions can be made (e.g. ranges of acceleration, velocity, and turn angles). While this may increase safety at first, it could, however, decrease performance. The performance could even be so low that due to new or strange behavior of the surrounding objects (e.g. like very risky take-over maneuvers) again safety is put at risk.

| Sub Challenge 4.3 |
|---|
| Handling the reactive nature of the interaction of a system with its surrounding objects. |

The prediction of the possible behavior of the surrounding objects has to be updated frequently because intentions of the objects can quickly change. Therefore, prediction needs to take into account not only the intention in the current situation, which is already hard to obtain, but also the influence of the predicting system on the predicted surrounding objects. For example, prediction is often used in the context of planning, here, the decision of an automated vehicle to avoid an obstacle left or right can reduce the available safe decision for other traffic participants and, in turn, force them to change their intention (cf. collision avoidance in aerospace, maritime).

| Sub Challenge 4.4 |
|---|
| Anticipating and handling changes in ODDs during the lifetime of the system. |

Even if all relevant objects in the ODD and their possible behavior could be captured precise and complete enough and highly automated or autonomous systems could predict the objects good enough, still the ODD could change, and the systems need to adapt. Such a change is even very likely because the introduction of automated systems (just like the introduction of e-scooters to traffic) will change the ODD because of shifts in the distributions of the relevant object's behavior. Other reasons for this kind of shifts may be changes in regulations (law, traffic rules), changes in characteristics (faster or more agile bikes), new objects, or changes in appearance (like clothes for sport or music events, or face masks during a pandemic). Such shifts can only be guessed a priori, and systems need to be adapted during their lifetime. However, too pessimistic approximation could again decrease performance to an unacceptable level that puts safety at risk.

| Sub Challenge 4.5 |
|---|
| Prediction of future security attacks and ensuring security of connected and evolving systems. |

Predicting behavior of relevant objects is already difficult regarding safety, but it is even harder for security in case of connected systems. Attack vectors are often unknown, and a combination of penetration testing and expert knowledge is needed to detect them. Attacks are surprising, one example is measuring timing of instructions of cryptographic algorithms when executed by CPUs that employ branch prediction. In the past, such measuring has been exploited to derive knowledge about the secret (cf. side-channel attacks such as Spectre).

Security is a prominent concern for communicating systems as they are subject to remote attacks. One way to cope with such attacks is to introduce components whose purpose is to detect the attacks. This imposes another challenge as such mechanisms open another attack surface, e.g. an attacker may force the system to run a certain mitigation strategy and as a result to reduce performance or even provoke a denial of service. For cloud-based automated systems this again puts safety at risk.

# Conclusion

Society has great expectations for a future with highly automated and autonomous systems. In this work we describe four main challenges which need to be solved in order to develop and validate safe, reliable and cyber-secure automated vehicles within a life cycle spanning design and test process and product operation. Each main challenge has been detailed in several sub-challenges, which go beyond current state-of-art approaches. Many aspects of the identified challenges also apply to the development of safe, secure and reliable ACPS in other areas (e.g., smart living, biomedical systems). Accepted and implementable standards and the incorporation of continuous engineering life-cycle processes within homologation/certification processes that are fit for purpose, accepted by regulation authorities and trusted by end-users are the long-term goals.

The discussion of these challenges shall support the objectives and orientation of further research, which we see of utmost importance to enable safe, reliable and secure automated vehicles for automated transportation and smart mobility, with convincing means of proof. Whoever addresses these challenges now and profoundly, will have a major impact on the future of highly automated vehicles, and consequently the best opportunities in a huge market that will emerge from it. Resulting methods and tools shall lead towards HAVs that provide sufficient confidence to allow future certification and earn sustainable trust of our society. The authors urgently recommend aligning upcoming research agendas according to the challenges identified, investing in the corresponding pre-competitive research at an early stage and promoting a strong industry transfer.

| Authors | Affiliation | Link |
|---|---|---|
| Nadja Marko | VIRTUAL VEHICLE | https://www.v2c2.at/ |
| Eike Möhlmann | OFFIS | https://www.offis.de/ |
| Dejan Nickovic | AIT Austrian Institute of Technology | https://www.ait.ac.at/ |
| Jürgen Niehaus | SafeTRANS | https://www.safetrans-de.org/ |
| Peter Priller | AVL | https://www.avl.com/ |
| Martijn Rooker | TTTech | https://www.tttech.com/ |

The authors would like to thank the following experts for their valuable comments on the manuscript: Jens Henkner, CertX AG; Sytze H. Kalisvaart, TNO; Philip Koopman, CMU; and Daniel Watzenig, TUG.

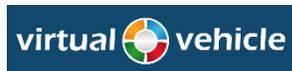
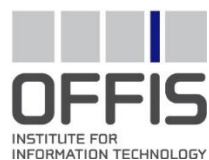
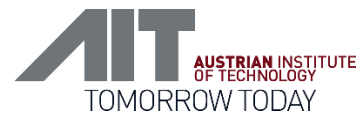
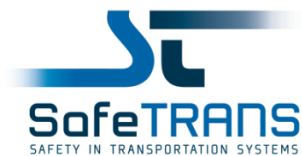
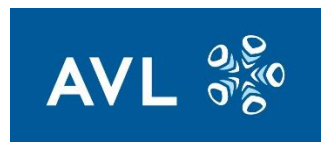
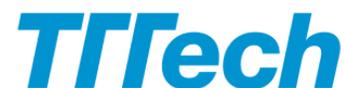